\documentclass[10pt,conference]{IEEEtran}

\usepackage[utf8]{inputenc}
\usepackage[portuguese,english]{babel}
\usepackage{cite}
\usepackage{graphicx}
\usepackage[cmex10]{amsmath}
\usepackage{amsthm}

\usepackage{algorithmic}
\usepackage{array}
\usepackage{booktabs}
\usepackage{todonotes}

\ifCLASSOPTIONcompsoc
  \usepackage[caption=false,font=normalsize,labelfont=sf,textfont=sf]{subfig}
\else
  \usepackage[caption=false,font=footnotesize]{subfig}
\fi
\usepackage{url}
\graphicspath{{img/}}

\begin{document}
\title{Identificação automática de pichação a partir de imagens urbanas}

\newif\iffinal
\finaltrue
\newcommand{\jemsid}{0000}

\iffinal

\author{\IEEEauthorblockN{Eric K. Tokuda and Roberto M. Cesar-Jr.}
\IEEEauthorblockA{Institute of Mathematics and Statistics\\
University of São Paulo (USP)\\
Brazil
}
\and
\IEEEauthorblockN{Claudio Silva}
\IEEEauthorblockA{Tandon School of Engineering\\
New York University (NYU)\\
USA
}
}

\else
  \author{SIBGRAPI paper ID: \jemsid \\ }
\fi
\maketitle

\begin{abstract}
	Graffiti tagging is a common issue in great cities an local authorities are on the move to combat it. The tagging map of a city can be a useful tool as it may help to clean-up highly saturated regions and discourage future acts in the neighbourhood and currently there is no way of getting a tagging map of a region in an automatic fashion and manual inspection or crowd participation are required. In this work, we describe a work in progress in creating an automatic way to get a tagging map of a city or region. It is based on the use of street view images and on the detection of graffiti tags in the images.
\end{abstract}

\selectlanguage{portuguese}
\begin{abstract}
	A pichação é um problema comum em grandes cidades e autoridades locais se esforçam para combatê-la. O mapa de pichação de uma região pode ser um recurso muito útil, pois pode auxiliar no combate ao vandalismo em regiões com alto nível de pichações e também a limpeza de regiões saturadas para desestimular atos futuros na mesma região. Atualmente não existe uma maneira automática de se obter o mapa de pichação de uma região e atualmente ele é obtido pela inspeção manual da polícia ou pela participação popular. Nesse sentido, descrevemos um trabalho em andamento no qual propomos uma forma automática de obter um mapa de graffiti de uma região geográfica. Ele se baseia no uso de imagens com vista de rua e na detecção de pichação nas imagens.
\end{abstract}

\IEEEpeerreviewmaketitle
\section{Introdução}

\begin{figure}[ht]
\centering
\begin{tabular}{cc}
\subfloat{\includegraphics[width=0.22\textwidth]{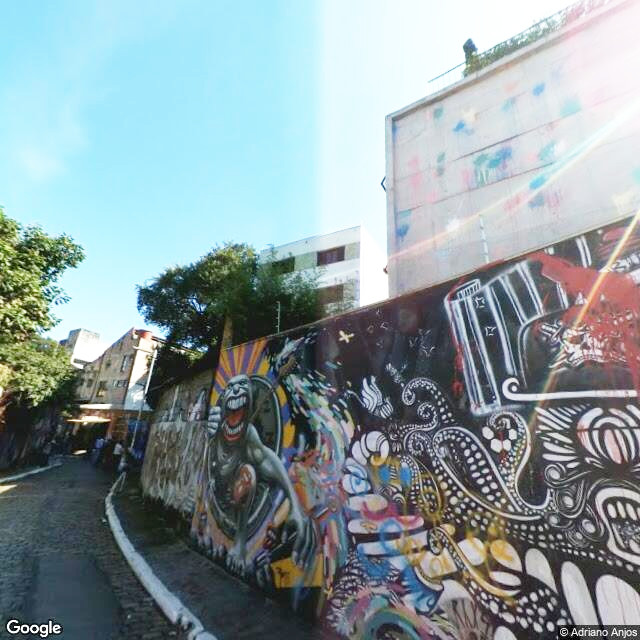}} &
\subfloat{\includegraphics[width=0.22\textwidth]{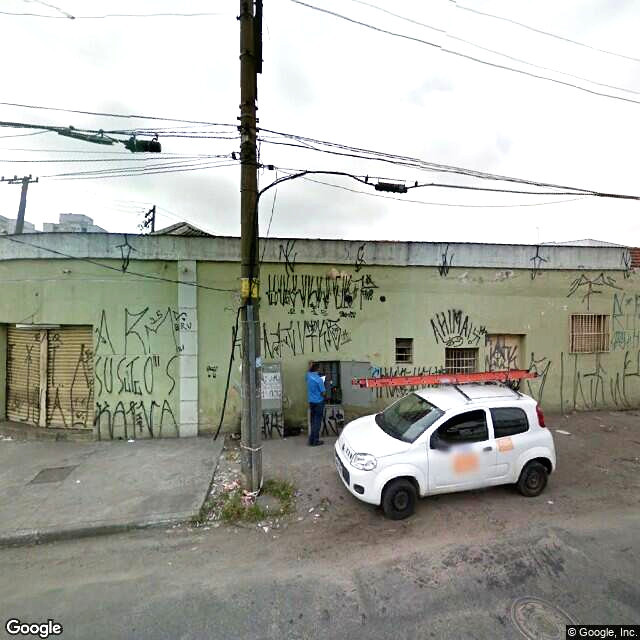}}
\end{tabular}
	\caption{Grafite e pichação. As pinturas de grafite são frutos de trabalhos meticulosos, enquanto as pichações são frequentemente atos não autorizados e compostos por letras e textos. Imagens obtidas a partir do serviço de visualização de ruas do~\cite{googlemaps}.}
\label{fig:grafftypes}
\end{figure}

Atualmente o grafite já faz parte do cenário das grandes cidades. Pode ser categorizado como grafite artístico ou pichação, como exemplificado na Figura~\ref{fig:grafftypes}, e enquanto o grafite é considerado uma expressão artística e, como tal, requer habilidades específicas, a pichação é geralmente um ato não autorizado que as pessoas simplesmente exibem frases ou nomes e a aceitação sobre pichação como arte é controversa~\cite{mcauliffe2012graffiti, young2013street}. Em 2017, a prefeitura de São Paulo, Brasil decretou uma lei que~\footnote{https://www1.folha.uol.com.br/cotidiano/2017/02/1860352-doria-sanciona-lei-anti-pichacao-e-veta-ate-grafite-nao-autorizado.shtml} que definia uma multa de até R\$$10.000,00$ aos autores de crimes de pichação contra patrimônios públicos.

Atualmente não existe uma modo automático de se criar o mapa de pichação de uma região geográfica e a criação por inspeção manual é uma tarefa dispendiosa. Neste trabalho descrevemos o trabalho em curso de uma metodologia para a criação de um mapa de pichação baseado na segmentação de regiões pichadas em imagens urbanas de público acesso.

\section{Trabalhos relacionados}

Trabalhos anteriores exploraram a tarefa de identificação de pichação em uma imagem~\cite{graffititracker,tagrs,graffititracking,vandaltrack} e paralelamente uma série de trabalhos utiliza dados geolocalizados para análises sociais, econômicas e culturais~\cite{doersch2012makes,zhou2014recognizing,arietta2014city}. Nenhum trabalho, porém, investigou a variação da concentração de pichação nos diferentes bairros da cidade e sua correlação com indicadores socio-econômicos e culturais.

Ferramentas de combate à pichação~\cite{graffititracker,tagrs,graffititracking,vandaltrack} utilizam participação colaborativa e permitem que usuários equipados com telefones celulares reportem atos de pichação. Alguns trabalhos~\cite{angiati2005novel,di2008graffiti,tombari2008graffiti} atacam o problema de uma maneira diferente e tentam identificar o \emph{ato} de pichar. Outros trabalhos recuperam pichações similares em um banco de dados de referência~\cite{yang2012efficient} usando componentes conexos e pontos-chave correspondentes em uma tentativa de associar pichação a gangues. Outra abordagem busca identificar a autoria do pichação~\cite{tong2011gang} dada uma imagem de teste através da recuperação de imagens similares, cálculo de uma métrica baseada nos símbolos contidos, anotação manual e correspondência entre os pontos-chave das imagens de pichação e as da gangue.

Como um sinal da relevância do tema em nível global, a União Européia tem um projeto~\cite{synyo2016graffolution} dedicado à análise dos principais atores envolvidos nos atos de pichação, incluindo escritores, cidadãos, autoridades policiais e administração pública. Este projeto inclui entrevistas com as partes interessadas e o estabelecimento de uma plataforma web que permitem a discussão e o compartilhamento de idéias sobre o tema de diferentes perspectivas. No Brasil, cidades altamente densas como São Paulo também enfrentam a presença generalizada de pichações na cidade~\cite{theguardian2016pixacao}.

A segmentação \emph{semântica} é uma tarefa de Visão Computacional que visa dividir a imagem em classes conhecidas~\cite{arbelaez2012semantic}. É uma tarefa complexa quando comparada com a classificação de imagens e a detecção de objetos, pois requer a classificação de cada pixel. A pesquisa nesta área é ativa e trabalhos recentes alcançam altos níveis de acurácia~\cite{arbelaez2012semantic, ronneberger2015u, badrinarayanan2015segnet, zhou2017scene, long2015fully}. Uma tarefa relacionada é a segmentação \emph{de  instâncias}, onde o objetivo é adicionalmente identificar as instâncias. Por exemplo, no caso de dois objetos com intersecção, o método deve ser capaz de identificar os limites de duas instâncias. Trabalhos anteriores~\cite{pinheiro2015learning, dai2016instance} atacaram o problema precedendo o estágio de detecção de objetos com um estágio de segmentação. Trabalhos de aprendizagem profunda (\emph{deep learning}) vêm atingindo os melhores resultados em diversas áreas da Visão Computacional~\cite{ronneberger2015u,hu2017finding,de2018exposing,silver2016mastering}, o que se observa também na tarefa de segmentação. O trabalho de Mask-RCNN~\cite{he2017mask}, aborda a tarefa de segmentação de instâncias executando as propostas de classificação e de segmentação de maneira paralela. \cite{he2017mask} baseia-se na arquitetura Faster-RCNN~\cite{ren2015faster}, mas com uma ramificação adicional para segmentação de instância.

Serviços como~\cite{googlemaps,mapillary} fornecem acesso público a imagens com vista de rua. As imagens são obtidas em diferentes localizações geográficas, períodos de aquisição e pontos de vista. Uma série de trabalhos já utilizaram esse tipo de imagem para fins de análise urbana~\cite{rundle2011using, torii2009google, tokuda2018anovel}. Os autores de~\cite{doersch2012makes} utilizam imagens de vista da rua para comparar os elementos arquitetônicos de diferentes regiões geográficas. O trabalho~\cite{li2015assessing} propõe a avaliação da vegetação urbana usando o mesmo tipo de imagens.

\section {Materiais e métodos}

Para estimar com confiança o nível de pichação em uma região geográfica, propomos uma métrica, o \emph {nível de pichação}, calculada utilizando a segmentação das áreas pichadas em imagens de vista de rua.

\subsection {Aquisição}
Uma região geográfica de interesse é inicialmente definida e as imagens dessa região são adquiridas. Idealmente, toda a região deve ser mapeada, mas devido a limitações da cobertura das imagens fornecidas e a restrições computacionais, apenas uma amostra é considerada. Existem diferentes maneiras de realizar a amostragem~\cite{stehman1999basic}, podendo ser classificadas em amostragem \emph{aleatória} e \emph{sistemática}. A amostragem aleatória remove o viés da seleção, mas não garante cobertura, diferentemente da sistemática que garante cobertura com precisão arbitrária, mas adiciona viés.

\begin{figure} [h]
	\centering
        \includegraphics [width = 0.11 \textwidth] {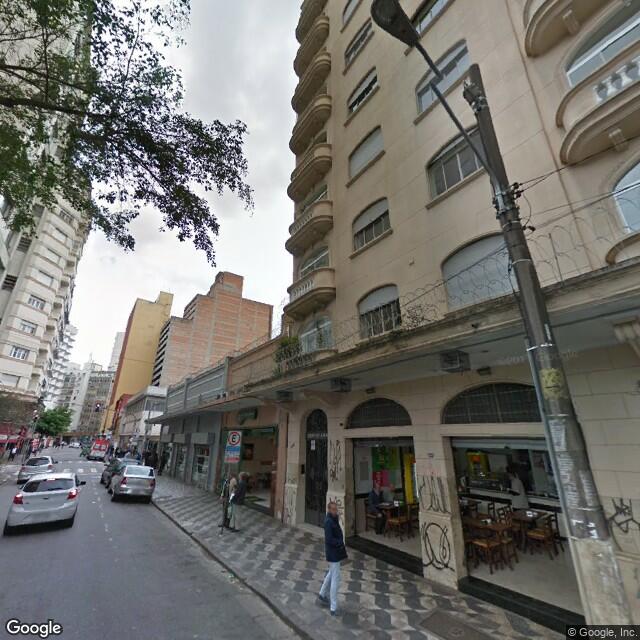}
        \includegraphics [width = 0.11 \textwidth] {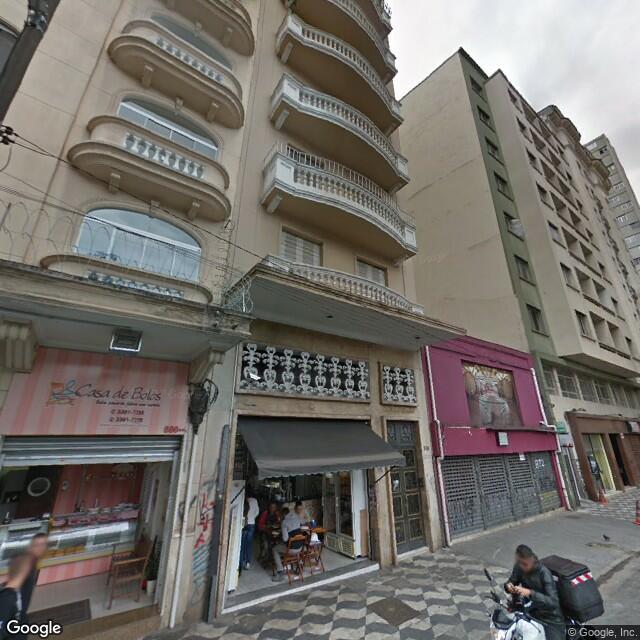}
        \includegraphics [width = 0.11 \textwidth] {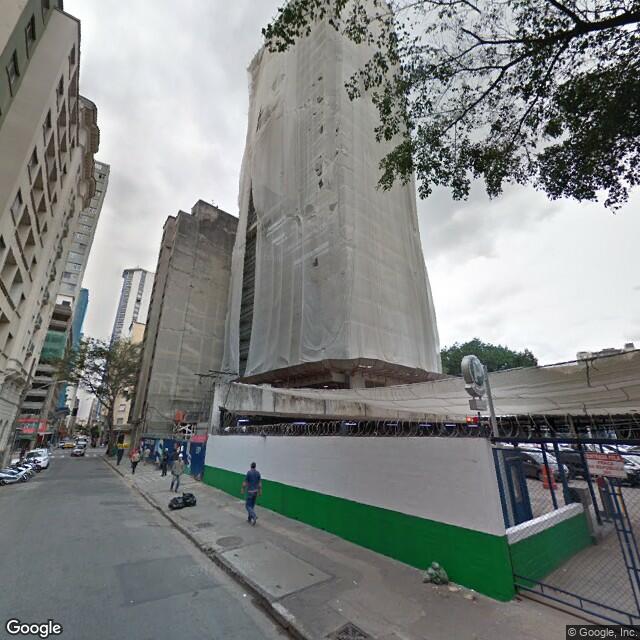}
        \includegraphics [width = 0.11 \textwidth] {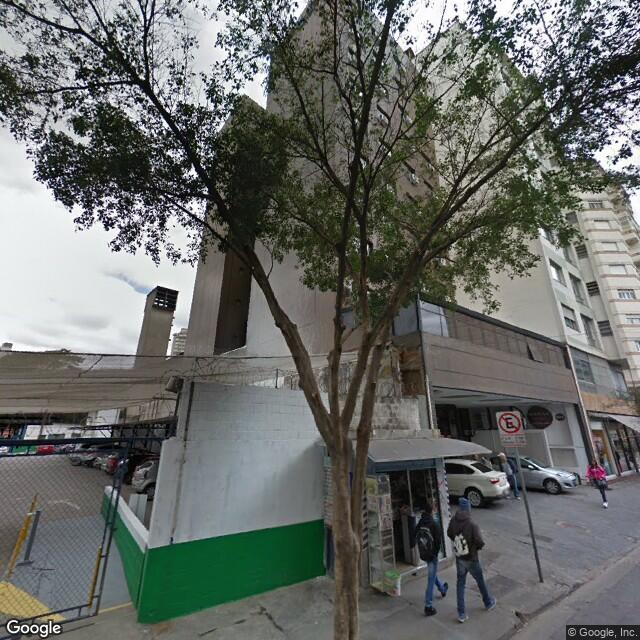}
\caption {Quatro visualizações da mesma localização geográfica. Imagens obtidas de~\cite{googlemaps}.}
        \label{fig:views}
\end{figure}

Uma vez que a amostra geográfica é definida, consideram-se as vistas de rua em cada ponto amostrado sendo que idealmente toda a cena visível em torno do observador deve ser considerada. Uma única vista panorâmica de $360^{\circ}$ pode ser usada, mas neste caso haverão  distorções presentes nas imagens obtidas. Alternativamente, pontos de vista complementares para cada local podem ser consideradas (veja a Figura~\ref{fig:views}).

\subsection {Identificação de pichação}

Dado o objetivo de quantificar o nível de pichação em um determinado local, uma maneira simples e direta seria classificar binariamente uma imagem se esta contém ou não pichação. No entanto, esta abordagem nos daria uma informação discreta e imprecisa de cada região e assim definimos o nível de pichação $ P(l) $ de uma localização geográfica $l$ como a soma das áreas pichadas $A$ em cada foto. Essa abordagem pode ser afetada pelo projeção e pela profundidade da cena. Assumimos que regiões distintas, dado uma amostra de tamanho mínimo, têm distribuição similar de projeções e de distância até os anteparos e, com essa suposição, $ P (l) $ pode ser usado para comparar regiões geográficas diferentes. Como representamos cada local $ l $ por um conjunto de $ k $ visões, definimos $ P(l)$ como a soma das áreas das regiões que contêm pichação em cada exibição. Podemos então agregar o nível de pichação por região geográfica e calcularmos a média dos níveis de pichação em nossa amostra de tamanho $ n $ (veja Equação~\ref{eq:perregion}).

\begin{equation}
P (R) = \frac {\sum_{j = 1}^{n} \sum_{i \in l_j}^{k} A_i} {n}
\label {eq:perregion}
\end{equation}

Nós optamos pelo método Mask-RCNN~\cite{he2017mask} para nossa tarefa de segmentação, dado seu alto desempenho relatado em importantes benchmarks~\cite{cordts2016cityscapes,lin2014microsoft}. Apesar de produzir segmentação e informação de instâncias, neste trabalho estamos apenas interessados na segmentação produzida.

Dada a inexistência de bases de dados de pichação disponíveis, criamos um conjunto de dados com imagens anotadas manualmente. Estes foram usados para treinar nosso modelo.

\section {Experimentos}

\begin {figure}[h]
\centering
\includegraphics [width = 0.4 \textwidth] {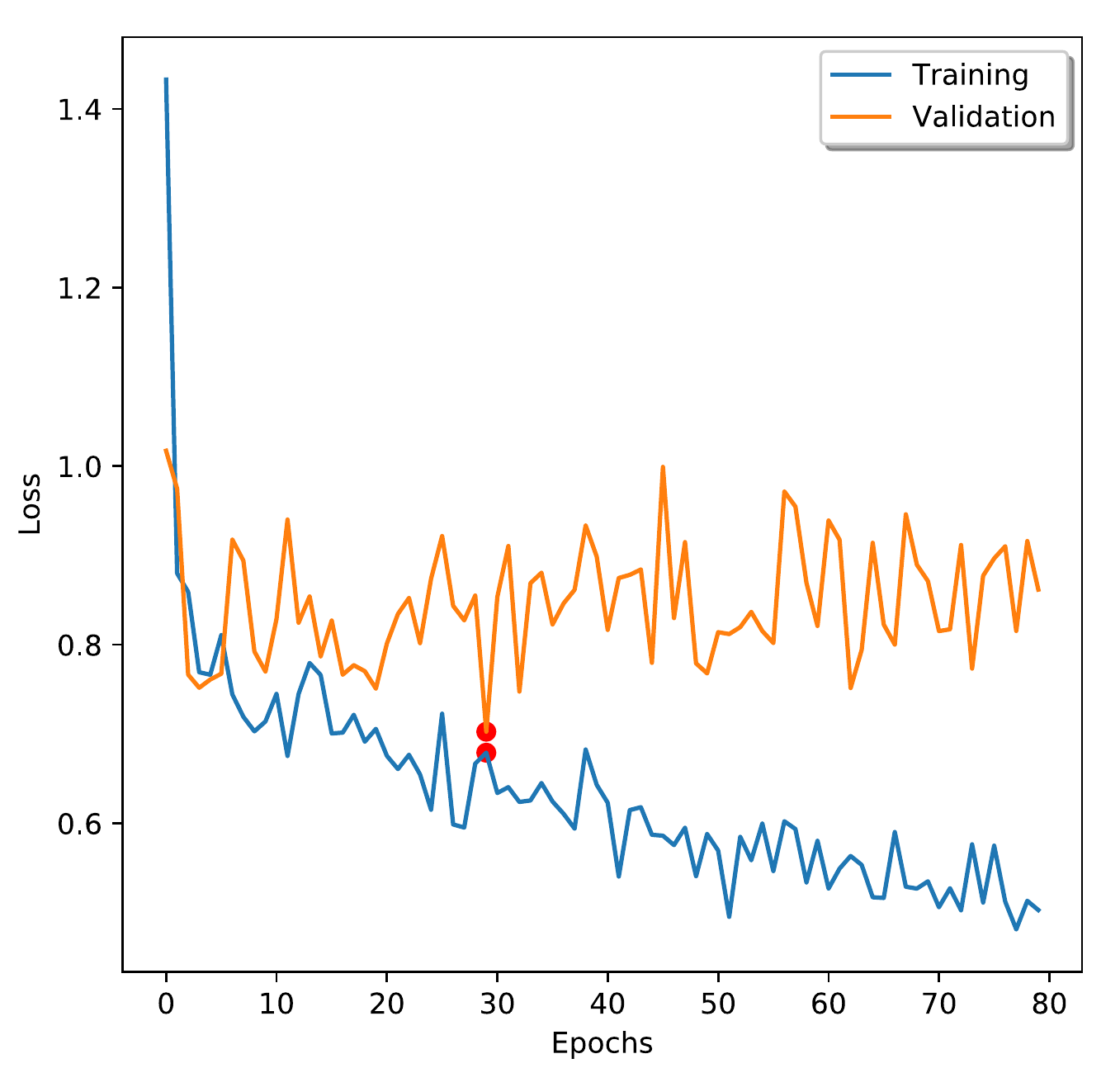}
\caption {Perda do modelo durante a etapa de treinamento.}
\label {fig:losses}
\end {figure}

\begin {figure*} [h]
\centering
\begin {tabular} {cccc}
\includegraphics [width = 0.29 \textwidth] {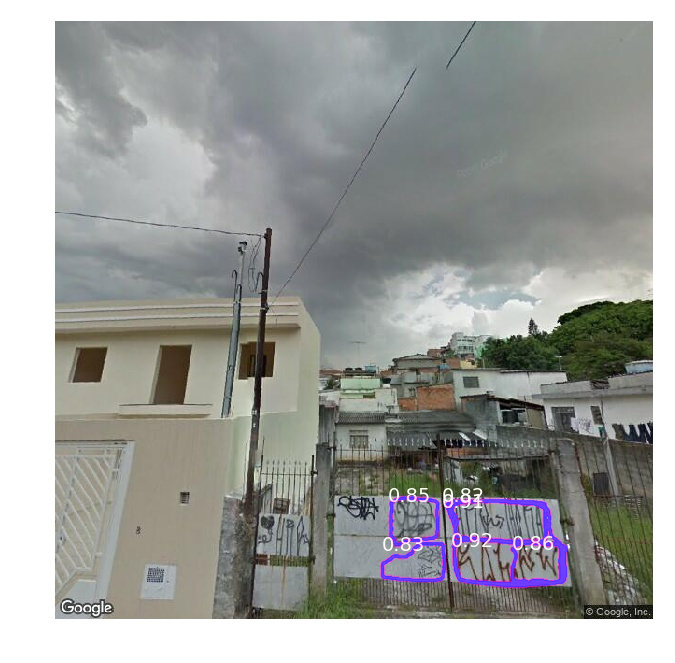} &
\includegraphics [width = 0.29 \textwidth] {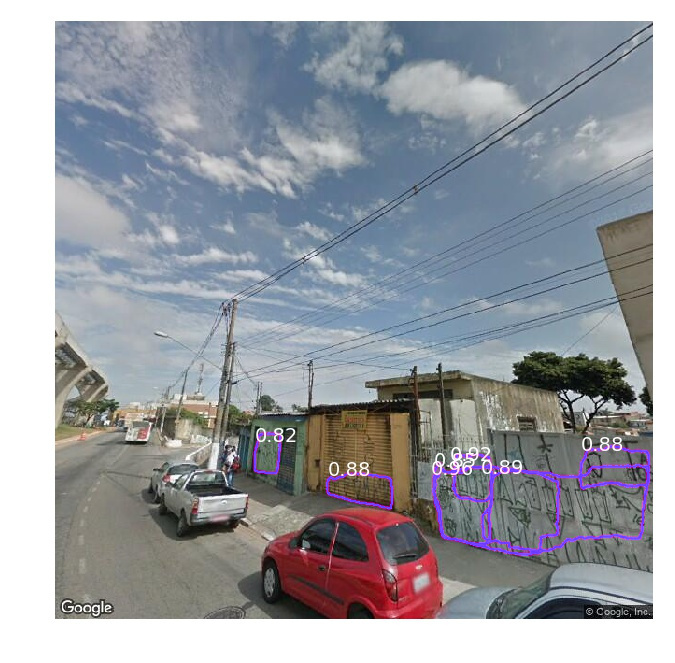} &
\includegraphics [width = 0.29 \textwidth] {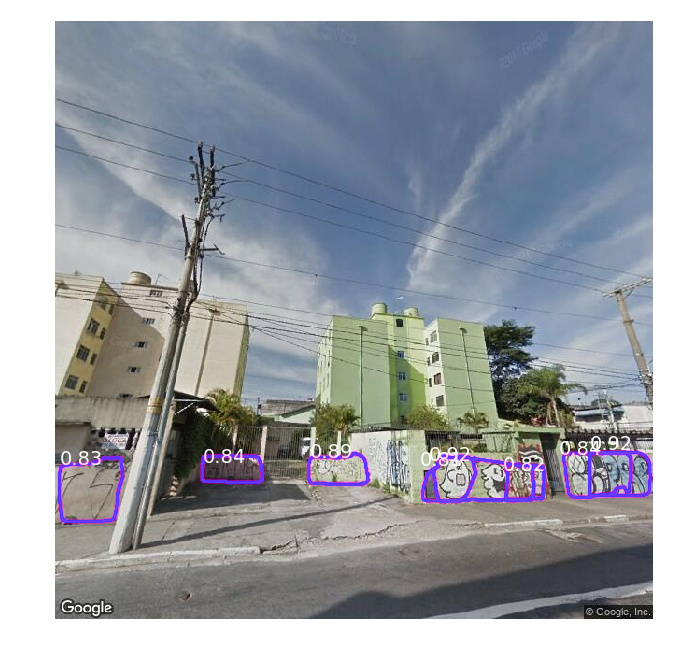} 
\end {tabular}
\caption {Amostra das detecções de pichação. Os valores descritos representam a confiança das detecções. Cenas provenientes de~\cite{googlemaps}.}
\label {fig:detections}
\end {figure*}

Inicialmente, coletamos uma amostra piloto de 10.000 imagens da cidade e as regiões que contêm pichação foram identificadas manualmente. Nosso conjunto de treinamento é composto de 632 imagens anotadas manualmente. Usamos uma arquitetura de redes residuais de 101 camadas~\cite {he2016deep} e um modelo pré-treinado no conjunto de dados COCO~ \cite {lin2014microsoft}. Utilizamos uma taxa de aprendizado de 0,001 e um momento de 0,9 e treinamos por 80 épocas. Usamos o modelo obtido na iteração de número 30, dado o seu menor erro de validação (ver Figura~\ref{fig:losses}). Utilizando a métrica de precisão médica proposto em VOC 2007 ~\cite{everingham2010pascal}, nosso modelo apresentou uma precisão média de $ 0,57 $. A Figura~\ref{fig:detections} apresenta uma amostra das detecções avaliadas. O tempo para o processamento de uma única imagem foi de $0,69$s em uma Geforce GTX 1050.

\begin {figure} [h]
\centering
        \includegraphics [width = 0.15 \textwidth] {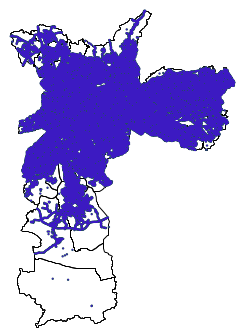}
\caption {Cobertura por imagens com vista de rua por~\cite{googlemaps} sobre a cidade de São Paulo.}
        \label {fig:coverage}
\end {figure}

Na Figura~\ref{fig:coverage} podemos ver a cobertura heterogênea das imagens fornecidas por~\cite{googlemaps}. Os dois distritos inferiores apresentaram pouca cobertura no momento de nossa aquisição, dada a natureza predominantemente rural e despovoada dessas regiões e, portanto, não foram considerados neste estudo.

Usamos quatro visualizações para cada localização geográfica, espaçadas por $90^{\circ} $. Observe na Figura~\ref{fig:views} como elementos de figuras adjacentes se cruzam, o que indica uma ampla cobertura de cada localização geográfica. $57\%$ das imagens consideradas são de 2017, como pode ser visto na Tabela~\ref{tab:years}.


\begin {table} [h]
\centering
\setlength \tabcolsep {2pt}

\caption {Ano de aquisição da amostra analisada}
\begin {tabular} {lccccccccc}
\toprule
\textbf {Year} & 2010 & 2011 & 2012 & 2013 & 2014 & 2015 & 2016 & 2017 & 2018 \\
\midrule

\textbf {Pontos} & 1,241 & 16,311 & 207 & 422 & 2,182 & 4,563 & 4,211 & 39,391 & 317 \\
\bottomrule
\end {tabular}
\label {tab:years}
\end {table}

Criamos uma malha sobre a extensão espacial da cidade com 134.624 pontos. Adotamos um espaçamento vertical e horizontal de 102 metros da nossa malha. Depois de eliminar pontos cujas imagens são de provedores externos e regiões não mapeadas (ver Figura~\ref{fig:coverage}), obtivemos uma cobertura geográfica de 68.752 pontos geográficos e 275.339 imagens no total.


\section{Considerações finais}

Este trabalho apresenta uma projeto em curso sobre o mapeamento automático de pichação em uma região geográfica. Utilizamos imagens de rua de uma região amostrada sistematicamente a partir da base do Google Maps~\cite{googlemaps} e identificamos as pichações em cada imagem. Propomos uma métrica para o nível de pichação de uma região geográfica.

Importante ressaltar que a métrica de pichação proposta é sensitiva à amostragem considerada, pois ela é calculada como uma média sobre os pontos \emph{amostrados}. O método proposto também é sensitivo a escolha da abordagem de segmentação utilizada, cuja acurácia impacta diretamente o resultado final. Etapas em andamento incluem o teste com outros algoritmos de segmentação e uma análise de pichação na cidade de São Paulo. Passos futuros incluem a utilização do método com uma amostragem mais densa, comparação de diferentes regiões geográficas e identificação de pichações recentes através da utilização de bases espaço-temporais~\cite{tokuda2018anew}.
~

\section*{Agradecimentos}

Os autores agradecem à Fundação de Amparo à Pesquisa do Estado de São Paulo, processos \#2014/24918-0, \#2015/22308-2 e ao CNPq, CAPES e NAP eScience - PRP - USP.

\bibliographystyle{IEEEtran}
\bibliography{references}
\end{document}